\documentclass{article}
\usepackage{microtype}
\usepackage{graphicx}
\usepackage{subfigure}
\usepackage{capt-of}
\usepackage{float} 
\usepackage{hyperref}
\usepackage{xurl} 
\usepackage{xcolor}

\usepackage[accepted]{mlsys2025}

\usepackage{amsmath,amsfonts,bm}

\def\eqref#1{equation~\ref{#1}}

\def\1{\bm{1}}

\DeclareMathAlphabet{\mathsfit}{\encodingdefault}{\sfdefault}{m}{sl}
\SetMathAlphabet{\mathsfit}{bold}{\encodingdefault}{\sfdefault}{bx}{n}

\hypersetup{pdfsubject={Jet-QSparse preprint}}
\makeatletter
\renewcommand{\mlsys@appearing}{\textit{Preprint.}}
\newenvironment{teaserfigure}{%
  \par\addvspace{0.1in}%
  \begin{minipage}{\textwidth}%
  \def\@captype{figure}%
  \centering
}{%
  \end{minipage}\par\vspace{0.1in}%
}
\makeatother

\mlsystitlerunning{Jet-QSparse: Accelerating Agentic LLM Inference with 4-Bit Sparse Attention}

\begin{document}

\twocolumn[
\mlsystitle{Jet-QSparse: Accelerating Agentic LLM Inference with 4-Bit Sparse Attention}

\begin{mlsysauthorlist}
\mlsysauthor{Haozhan Tang}{cmu}
\mlsysauthor{Hao Kang}{cmu}
\mlsysauthor{Meng Lu}{vt}
\mlsysauthor{Yecheng Wu}{mit}
\mlsysauthor{Zerui Wang}{nvidia}
\mlsysauthor{Seokho Han}{skku}
\mlsysauthor{Haocheng Xi}{berkeley}
\mlsysauthor{Shuo Yang}{berkeley}
\mlsysauthor{Song Han}{mit,nvidia}
\mlsysauthor{Han Cai}{nvidia}
\end{mlsysauthorlist}

\mlsysaffiliation{cmu}{Carnegie Mellon University}
\mlsysaffiliation{vt}{Virginia Tech}
\mlsysaffiliation{mit}{Massachusetts Institute of Technology}
\mlsysaffiliation{nvidia}{NVIDIA}
\mlsysaffiliation{skku}{Sungkyunkwan University}
\mlsysaffiliation{berkeley}{University of California, Berkeley}

\mlsyscorrespondingauthor{Han Cai}{hcai@nvidia.com}

\vskip 0.3in

\begin{abstract}
Long-horizon reasoning and agentic workloads strain KV-cache capacity and decode attention throughput.
We present Jet-QSparse, a training-free method that combines compact NVFP4 cache storage with native FP4 query--key computation on Blackwell tensor cores. 
A dense scan scores the full key history and selects 128-token blocks for sparse FP16 value aggregation with FP32 accumulation, preserving higher precision in the output-forming step while enabling contiguous tile loads.
Calibrated key centering and at most two BF16 pages per sequence and layer address quantization error; FP8 prefill uses a temporary arena while the persistent cache is packed directly from BF16 activations. 
Across four hybrid LLMs, a budget of 128 blocks per KV head shows no detected accuracy loss on the evaluated mathematical reasoning, coding, and agentic benchmarks, although some smaller budgets incur losses.
With 64 blocks, Qwen3.8-27B achieves a $1.94\times$ end-to-end decode speedup over dense BF16 at 128K context and batch size 16 on one GB300, and the measured Qwen KV pools provide $2.58$--$2.76\times$ BF16's \mbox{token capacity}.
\end{abstract}
\begin{teaserfigure}
\includegraphics{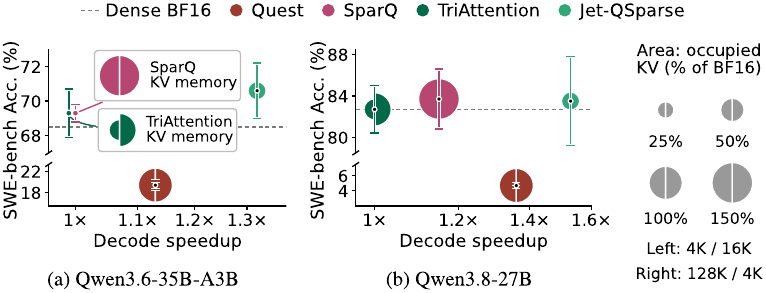}
\captionof{figure}{%
SWE-bench Verified accuracy (mean $\pm$ SD, three seeds) versus separate
GB300 decode speedups (128K prompt/4K output, batch 8); Jet-QSparse uses $N=64$.
Dashed lines mark BF16 accuracy. Leaders connect memory callouts to measured points.
Semicircle areas are proportional to occupied KV (\% of BF16):
left, 4K/16K (one run); right, 128K/4K (two runs), both at batch 8.
Panels share one area scale; x-axis ranges differ.
Bytes include pages, metadata, and active BF16 copies,
excluding unused reservations, recurrent state, and prefill workspace.}
\label{fig:teaser}
\end{teaserfigure}

]

\printAffiliationsAndNotice{}

\section{Introduction}
\label{sec:introduction}

Reasoning and agentic language models process long histories of intermediate
reasoning, tool outputs, and interaction. The key--value (KV) cache grows with
these histories and limits the number of requests that fit on a GPU. Dense
decode attention reads and processes the history for every generated token,
increasing latency as the sequence grows. Efficient inference
requires a smaller persistent cache and cheaper attention while
preserving task accuracy.

KV quantization reduces the bytes stored per token
\citep{liu2024kivi,hooper2024kvquant}, while sparse decode reduces the data
fetched and processed for each query
\citep{tang2024quest,ribar2024sparq,yuan2026blasst}.
Blackwell supports native NVFP4 computation: packed 4-bit operands reduce memory traffic and
feed block-scaled tensor-core operations \citep{nvidia2025nvfp4,du2026bitdecoding}.
An efficient attention method must coordinate the precision used to score
keys, the blocks selected for value aggregation, and the layout used to
fetch them. Selection
adds work, irregular accesses can limit hardware utilization, and errors
from quantization and sparsification must be evaluated together.

\textbf{Jet-QSparse} combines compact NVFP4 storage and dense native FP4
scoring with sparse higher-precision aggregation in a simple, training-free
design. Its NVFP4 cache retains the complete token history.
At each decode step, native FP4 query--key computation scores every key
and estimates block attention masses. Query-dependent selection restricts
the output-forming $PV$ product to selected blocks, using FP16 operands and FP32
accumulation. Calibrated key centering and BF16 copies of the first and
current blocks address quantization error within two pages per
sequence and layer.

We group 128 consecutive tokens per cache page to align selection with
contiguous memory loads and tensor-core tiles. Query heads sharing a KV head
reuse selected blocks, and selection executes on the GPU within CUDA graphs.
Prepacked scales feed native QK computation, and pure decode steps fuse
query packing with cache writes.
The implementation integrates into a production serving engine. Prefill uses
FP8 attention while the persistent NVFP4 cache is packed directly from BF16
activations.

We study sparsification under quantization through attention-output analysis,
dense NVFP4 controls, and block-budget sweeps. The analysis relates selection
error to omitted mass and the gap between selected and omitted value averages.
Long-context diagnostics show that aggressive sparsity can impair aggregation
over distributed evidence while largely preserving needle retrieval;
agentic tests reveal model-dependent failure patterns. These findings motivate
evaluating precision and sparsity together at the task level.

We evaluate Jet-QSparse on Qwen3.6-35B-A3B, Qwen3.8-27B, Nemotron 3
Nano, and Nemotron 3.5 Lightning. These hybrid models span different
generations, dense or mixture-of-experts (MoE) feedforward layers, and
different proportions of full-attention layers. We also deploy
Qwen3.8-2.4T-A95B across eight nodes. On the two Qwens, the measured cache
pools provide $2.58$--$2.76\times$ the KV-token capacity of dense BF16 under
the same allocation policy. At 128K context and batch size eight on one GB300,
64 selected blocks per KV head accelerate Qwen3.8-27B and Qwen3.6-35B-A3B
decode by $1.53\times$ and $1.32\times$ over dense BF16, respectively.
Every compared method fits this workload. Both configurations show no detected accuracy loss on five
reasoning, coding, and agentic benchmarks at the evaluation's statistical
resolution; a 128-block budget meets this criterion across all four models.
Figure~\ref{fig:teaser} compares speed, accuracy, and occupied KV memory
at this shared budget.
Smaller budgets incur losses on some model--benchmark pairs, which we
report alongside the performance gains.

Our contributions are:
\begin{enumerate}
  \item We introduce Jet-QSparse, a simple, training-free attention method
  that combines compact NVFP4 storage and native low-precision scoring with
  sparse higher-precision aggregation. Its block-based cache aligns selection
  with contiguous memory access and tensor-core execution, enabling faster
  decoding and lower KV memory use within a production serving engine.
  \item We systematically study the additional error from sparsifying
  quantized attention and its effects on long-context and agentic tasks.
  We relate selection error to omitted mass and value differences, and use
  Jet-QSparse's dense NVFP4 controls and budget sweeps to examine task
  sensitivity and model-dependent \mbox{failure patterns}.
  \item We evaluate Jet-QSparse across Qwen and Nemotron model families on
  mathematical reasoning, coding, and agentic benchmarks, demonstrating
  decode speedups and increased KV capacity. A 128-block budget shows no
  detected accuracy loss on the five application benchmarks across all four
  models, and we extend the evaluation to a 2.4T-parameter model with 95B
  \mbox{active parameters}.
\end{enumerate}

\section{Related Work}
\label{sec:related-work}

\paragraph{Hybrid Attention Models.}
Linear attention uses fixed-size states for recurrent decoding
\citep{katharopoulos2020linearattention}.
Mamba uses selective state-space recurrences; Mamba-2 relates them to
structured attention through state-space duality
\citep{gu2023mamba,dao2024mamba2}.
Gated DeltaNet (GDN) combines state-decay gates with delta-rule updates
to key--value associations \citep{yang2025gateddelta}.
Hybrids interleave these recurrent mechanisms with softmax attention.
Qwen3-Next-80B-A3B, Qwen3.6-35B-A3B, and Qwen3.8-27B each interleave
three GDN blocks with one gated full-attention block
\citep{qwen2025next,qwen2026qwen36,qwen2026qwen38}.
Nemotron 3 Nano and Nemotron 3.5 Lightning combine Mamba-2, MoE, and
six full-attention GQA blocks in 52-block backbones
\citep{nvidia2025nemotron3nano,nvidia2026nemotron35}.
Jet-QSparse reduces the KV-cache costs of these attention layers
without changing recurrent or feedforward blocks.

\paragraph{Attention kernels and serving systems.}
FlashAttention-2/3/4 improve dense attention through tiling, work partitioning,
and hardware-specific pipelines
\citep{dao2024flashattention2,shah2024flashattention3,zadouri2026flashattention4}.
PagedAttention and FlashInfer provide cache management and paged
execution \citep{kwon2023pagedattention,ye2025flashinfer}.
TensorRT-LLM integrates paged execution, quantized caches, and multiple
attention backends \citep{trtllm_attention,trtllm_quantization}.
Its \texttt{trtllm-gen} backend pairs NVFP4 KV with FP8 queries
\citep{trtllm_fmha_backend}; Jet-QSparse uses NVFP4 Q and K in its dense scan.

\paragraph{Low-bit arithmetic and KV storage.}
SageAttention uses key centering and INT8 QK; SageAttention2 combines INT4
QK with FP8 PV; SageAttention3 extends both products to NVFP4
\citep{zhang2025sageattention,zhang2024sageattention2,zhang2025sageattention3}.
Attn-QAT studies quantization-aware training and provides kernels with
NVFP4 QK and higher-precision PV; Hardware-Aware FP4 FlashAttention-4 quantizes
probabilities for noncausal forward attention
\citep{zhang2026attnqat,hu2026hardwarefp4}.
KIVI quantizes keys per channel and values per token with a full-precision
residual buffer; KVQuant combines pre-RoPE key
quantization with nonuniform codebooks and separately stored outliers;
QuaRot uses invariant rotations
\citep{liu2024kivi,hooper2024kvquant,ashkboos2024quarot}.
BitDecoding maps dense low-bit-cache attention to tensor cores, including
native MXFP4/NVFP4 computation on Blackwell~\citep{du2026bitdecoding}.

\paragraph{Query-dependent sparse decode.}
Quest bounds page scores using channel-wise key minima and maxima
\citep{tang2024quest}.
SparQ estimates token importance from query coordinates with large magnitudes,
fetches selected full K/V vectors, and compensates for estimated omitted mass
using a running value mean \citep{ribar2024sparq}.
Loki uses an offline per-head PCA basis to rank tokens in reduced dimensions
before attending to selected full-dimensional vectors \citep{singhania2024loki}.
Jet-QSparse scans all NVFP4 keys and selects blocks by globally normalized
attention mass.

\paragraph{KV eviction and retention.}
H$_2$O keeps recent tokens and attention heavy hitters; StreamingLLM keeps
sinks and a rolling window; SnapKV compresses the prompt cache using a
trailing observation window
\citep{zhang2023h2o,xiao2024streamingllm,li2024snapkv}.
DuoAttention separates full-history retrieval and streaming heads through
optimized gates \citep{xiao2025duoattention}, while RocketKV
combines prompt eviction with per-step selection
\citep{behnam2025rocketkv}.
TriAttention periodically evicts generation KV using calibrated pre-RoPE
query statistics and trigonometric analysis \citep{mao2026triattention}.
ThinKV combines thought-dependent precision with progressive eviction
\citep{ramachandran2026thinkv}.
Jet-QSparse retains every token in NVFP4 and selects aggregation blocks
at each step.

\paragraph{Sparse prefill and trained selectors.}
MInference, FlexPrefill, and XAttention use training-free sparse prefill
patterns to reduce QK and PV work
\citep{jiang2024minference,lai2025flexprefill,xu2025xattention};
PAROAttention combines block sparsity and low-bit arithmetic for visual
generation \citep{zhao2025paroattention}.
SeerAttention trains block gates, and SeerAttention-R extends learned,
GQA-shared selection to reasoning decode
\citep{gao2025seerattention,gao2026seerattentionr}.
MoBA uses parameter-free routing in sparsely trained models;
Native Sparse Attention and DeepSeek Sparse Attention introduce learned
compression or indexing
\citep{lu2025moba,yuan2025nsa,deepseek2025v32}.
Jet-QSparse calibrates existing full-attention layers without weight updates.

\paragraph{Sparse aggregation after scoring.}
SpargeAttention combines pooled Q/K selection with online-softmax PV
skipping \citep{zhang2025spargeattn}.
BLASST compares tile and running maxima to skip softmax, V loads, and PV
without a separate selection pass \citep{yuan2026blasst}.
Jet-QSparse uses a separate dense NVFP4 scan to select shared 128-token
blocks under a fixed budget or capped tail-mass policy, then performs
FP16 PV with FP32 accumulation. Caps may override the mass target.

\section{Jet-QSparse: Hardware-Optimized 4-Bit Sparse Attention}
\label{sec:method}
\newcommand{\JQSOverviewCaption}{%
Jet-QSparse decode. Logical blocks pair key and value pages for 128 consecutive
tokens; internal divisions are schematic. The dense scan emits FP32 statistics
from every key block. First/current-block statistics are recomputed from BF16
keys before normalization. Each query-head group~\citep{ainslie2023gqa} shares selected K/V blocks,
always including the current block. Selected scores are recomputed for online
softmax. FP16 unnormalized weights multiply reconstructed FP16 values with FP32 accumulation;
denominator weights remain FP32. Merged partials are normalized and cast to BF16.
BF16 paths use the same quantized query and FP16 operands. Reconstruction retains
NVFP4 storage error. An unselected first page may still be read and masked.}

\newcommand{\JQSCacheCaption}{%
Blocked cache layout and lifetime. (a) K/V pages cover matching 128-token ranges;
logical neighbors need not be physically adjacent. (b) Bytes cover one head's
contiguous K or V region; compression excludes fixed buffers. (c) Token 384 opens
block 3, reusing the current BF16 slot. Block 0's BF16 copy~\citep{xiao2024streamingllm,hooper2024kvquant} and all NVFP4 copies persist;
the live row is enlarged. (d) BF16 inputs feed both paths independently. FP8 pages
become reusable; the arena remains reserved.}

\newcommand{\JQSSelectionCaption}{%
Illustrative block attention for two query heads. Green outlines mark selected
blocks. Fixed $N=3$ includes current block 7; counts are exact, ranks approximate.
Tail selection unions the head sets and current block, yielding six blocks with
nonbinding four-block caps. Binding caps can override the mass target.
Omitted percentages follow head order.}

Jet-QSparse uses NVFP4 as both a compact storage format and a native
tensor-core operand. Storing the KV cache in NVFP4 reduces its footprint and
read traffic; computing $QK^\top$ directly from packed queries and keys uses
Blackwell's FP4 tensor cores to score the full context. We retain FP16 operands
and FP32 accumulation for $PV$, the weighted sum that forms the attention
output. Sparsity reduces the cost of this higher-precision step by restricting
aggregation to selected 128-token blocks.

\begin{figure*}[t]
  \centering
  \includegraphics[width=\textwidth]{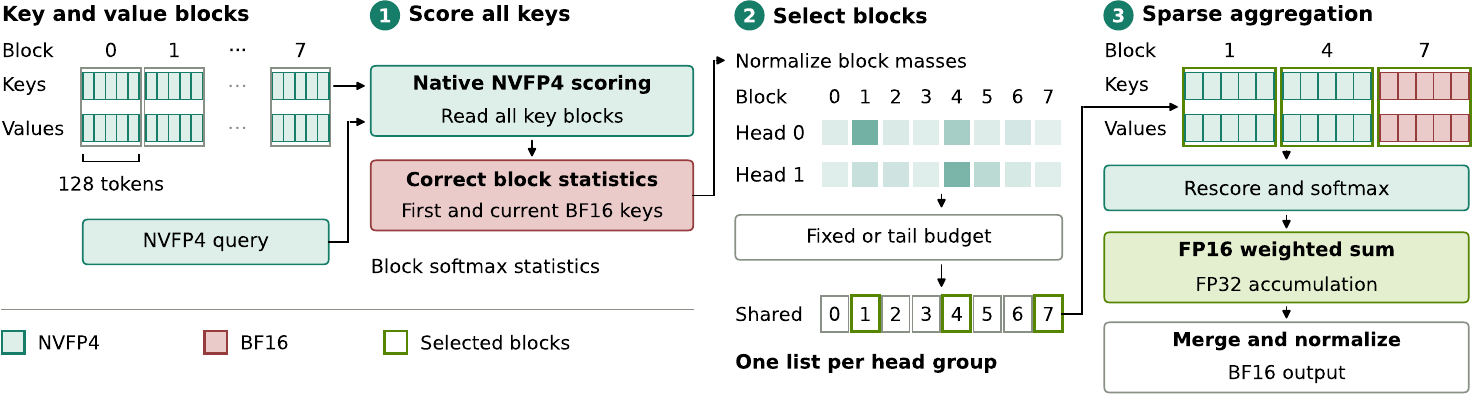}
  \caption{\JQSOverviewCaption}
  \label{fig:jqs-overview}
\end{figure*}

Figure~\ref{fig:jqs-overview} summarizes the pipeline: \emph{dense NVFP4 scoring
followed by sparse higher-precision aggregation}. Blocks align selection with
contiguous memory loads and tensor-core tiles. Calibrated key centering and BF16 copies of the
first and current blocks address quantization error within a two-page
allowance. The method targets single-token causal decode in full-attention
layers and requires no model-weight updates. Prefill uses a separate FP8 path
(Section~\ref{sec:method-integration}).

\subsection{NVFP4 Cache and Block Layout}
\label{sec:method-cache}

\begin{figure}[!t]
  \centering
  \includegraphics[width=\columnwidth]{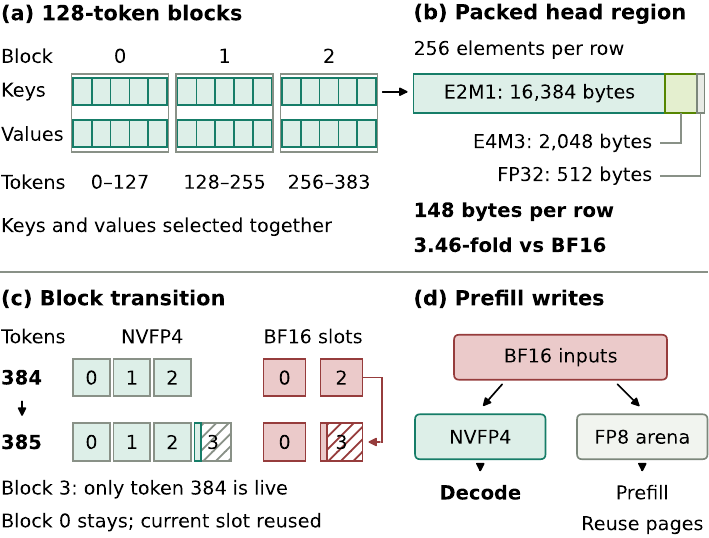}
  \caption{\JQSCacheCaption}
  \label{fig:jqs-cache}
\end{figure}

\paragraph{Aligning selection with memory access.}
Token-level selection can scatter memory accesses and create irregular
workloads. We group $B=128$ consecutive tokens into a block, storing its
keys and values in matching cache pages (Figure~\ref{fig:jqs-cache}).
We select whole token blocks. Each KV head occupies
contiguous regions for packed codes, micro-block scales, and row scales within
a page. A page table locates these regions so kernels can load contiguous
tiles even when selected pages are not adjacent in physical memory.
Using whole blocks trades finer selection granularity for regular
memory access and block-sized matrix products.

\paragraph{Packing keys and values.}
A row contains $d$ elements for one token and one KV head. Its NVFP4
reconstruction is $\hat{x}_i=a_x s_{x,\beta(i)}c_{x,i}$, where $c_{x,i}$ is an
E2M1 code, $s_{x,\beta(i)}$ is an E4M3 scale shared by a micro-block of 16
elements, and $a_x$ is an FP32 row scale. For zero-based element index $i$,
$\beta(i)=\lfloor i/16\rfloor$. Scaling the row's maximum magnitude by
$1/(6\cdot448)$ gives $a_x$, where 6 and 448 are the E2M1 and E4M3 maxima.
Each micro-block's maximum magnitude then sets its E4M3 scale, and its
elements are rounded to E2M1.
Zero scales are replaced by one, so an all-zero row reconstructs as zero.
Keys and values are packed independently; the query uses the same format.
Including scale metadata, a row occupies
\begin{equation}
  w_d = \frac{d}{2}+\frac{d}{16}+4 \quad \text{bytes}.
  \label{eq:jqs-row-bytes}
\end{equation}
This is 4.625 bits per element at $d=256$ and 4.75 at $d=128$:
$3.46\times$ and $3.37\times$ packed-row compression relative to BF16,
excluding fixed buffers.

\paragraph{Centering keys.}
A component shared across keys can consume quantization range without
affecting their relative attention scores. We subtract a calibrated mean
$\mu$ for each layer and KV head after applying RoPE, where used. In exact
arithmetic, subtracting this common vector adds the same constant to every
logit of a query and leaves softmax unchanged~\citep{zhang2025sageattention}.
Quantizing the centered keys
can still change attention. An offline attention-output error screen
determines which heads use a nonzero mean;
the remaining heads use $\mu=0$.

\paragraph{Retaining two BF16 pages.}
We retain BF16 copies of block 0 and the current block alongside their NVFP4
copies. The side buffer holds centered keys rounded to BF16 and the original
BF16 values, covering at most two distinct blocks per sequence and layer.
We pack every row into NVFP4 on its first write. After the current page
advances, we read a completed noninitial block from its existing NVFP4 copy;
block 0's BF16 copy persists throughout the sequence. We mask unused
positions in the current page. Selection determines whether block 0 contributes
to the sparse output.

\subsection{Dense Scoring for Block Selection}
\label{sec:method-scores}

Selecting blocks by their share of attention requires a normalizer over the
full context. We obtain it with a dense NVFP4 scan that reads keys and scales
without fetching values or materializing the attention matrix.

Consider one KV head and its group $\mathcal{G}$ of query heads. Let $L$ be
the cached sequence length, including the current token,
$n_b=\lceil L/B\rceil$ the number of live blocks, and $\mathcal{I}_b$ the live
token positions in block $b$. Denote the reconstructed NVFP4 query by
$\hat q_h$. The effective operands $k_j^{\mathrm e},v_j^{\mathrm e}$ are the
BF16 copies in the first and current blocks and the reconstructed NVFP4
operands elsewhere. In exact arithmetic,
\begin{equation}
\begin{aligned}
  z_{hj} &= \frac{\hat q_h^\top k_j^{\mathrm e}}{\sqrt d},
  &p_{hj} &= \frac{\exp(z_{hj})}{\sum_{t<L}\exp(z_{ht})},\\
  P_{hb} &= \sum_{j\in\mathcal{I}_b}p_{hj}.
\end{aligned}
\label{eq:jqs-block-mass}
\end{equation}
Thus, $P_{hb}$ measures a block's importance under the method's dense
attention distribution. Dense here refers to coverage of all keys.

The scan records a reference maximum and an exponential sum for each block
and query head. A correction kernel replaces the first and current blocks'
statistics using their BF16 keys and the same quantized query. After this
correction, we rescale the statistics to a common maximum and normalize
across blocks.
This gives selection and aggregation the same effective operands. We write
$\bar P_{hb}$ for the kernel's FP32 estimates of the ideal masses in
Equation~\ref{eq:jqs-block-mass}.

\subsection{Selecting Blocks for Grouped-Query Attention}
\label{sec:method-selection}

\begin{figure}[t]
  \centering
  \includegraphics[width=\columnwidth]{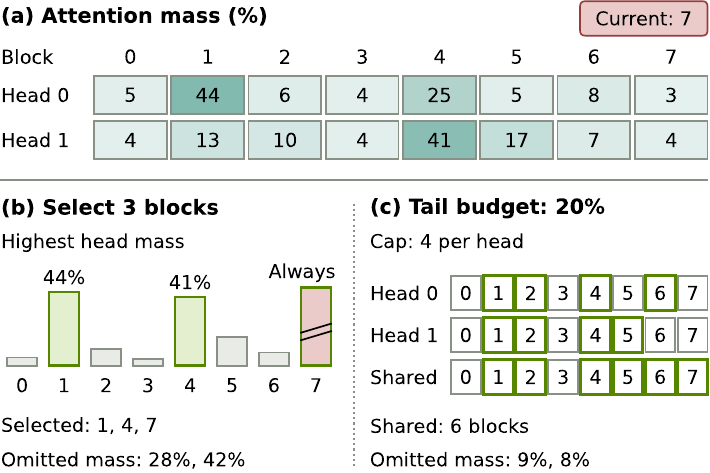}
  \caption{\JQSSelectionCaption}
  \label{fig:jqs-selection}
\end{figure}

Each KV head uses one selected block set $U$ for all query heads in its group,
sharing the fetched K/V tiles. A fixed budget controls aggregation work,
while a tail-mass policy adapts the selected count to the attention distribution.
Both policies include the current block $b_\star=n_b-1$.
Figure~\ref{fig:jqs-selection} illustrates their selections for one query-head group.

\paragraph{Fixed block budget.}
We rank each block by the largest mass assigned to it within the group,
$\pi_b=\max_{h\in\mathcal{G}}\bar P_{hb}$. A block can therefore rank highly
when any head assigns it substantial mass. Setting $\pi_{b_\star}=2$ ensures
that the current block occupies one of the $n=\min(N,n_b)$ selected slots.
A 16-step threshold bisection finds the ranking boundary; any remaining
slots are filled from the final boundary band in ascending block order.
The selected count is exact, while the top-$N$ ranking is approximate.

We evaluate $N=32,64,128$, selecting at most 4,096, 8,192, or 16,384 token
positions per KV head. All blocks are selected when the context fits within
the budget. A fixed count does not bound an individual head's omitted mass:
blocks preferred by other heads can fill the shared budget.

\paragraph{Tail-mass budget.}
Each query head $h$ has an omitted-mass budget $\tau_h$ and a calibrated cap
fraction $f_h$. We set $f_h$ to the 99th percentile of the context fraction
needed on held-out prompts at omitted mass 0.125. Calibration uses 16-token
pages; decode converts the fraction to a cap on 128-token blocks,
$c_h=\min(n_b,\max(1,\lceil f_hL/B\rceil))$, ignoring FP32 rounding in this
expression.

The tail threshold
$\theta_h^\tau$ keeps the sum of masses below it within $\tau_h$; the cap
threshold $\theta_h^c$ retains at most $c_h$ blocks. We find both by
16-step bisection on $[0,1]$, taking the feasible lower endpoint for the
tail constraint and the feasible upper endpoint for the cap:
\begin{equation}
\begin{aligned}
  \theta_h &= \max(\theta_h^\tau,\theta_h^c),
  & K_h &= \{b:\bar P_{hb}\ge\theta_h\},\\
  U &= \{b_\star\}\cup\bigcup_{h\in\mathcal{G}}K_h.
\end{aligned}
\label{eq:jqs-tail-selection}
\end{equation}
The cap takes precedence over the mass budget and applies to $K_h$, not the
final group union. Thresholding can underfill a head's cap, even leaving an
empty set. In the evaluated configurations, $\tau_h=1-\rho$ for every
head: $\rho=0.40$ and $0.60$ correspond to omitted-mass budgets of 0.60 and
0.40. Only $f_h$ varies across heads. $\rho$ sets a mass target rather than
a fixed block fraction.

\subsection{Sparse Value Aggregation}
\label{sec:method-aggregation}

We fetch the selected pages and recompute their QK products, avoiding a
write and reread of token-level scores from the dense scan. With
$\mathcal{K}(U)=\bigcup_{b\in U}\mathcal{I}_b$, the ideal output is
\begin{equation}
  o_h^U =
  \frac{\sum_{j\in\mathcal{K}(U)}p_{hj}v_j^{\mathrm e}}
       {\sum_{j\in\mathcal{K}(U)}p_{hj}}.
  \label{eq:jqs-selected-output}
\end{equation}
This renormalizes softmax over selected keys only.

\paragraph{Effect of omitted mass.}
Let $o_h^{\mathrm D}$ be dense attention on the same effective operands and
$\varepsilon_h=\sum_{j\notin\mathcal{K}(U)}p_{hj}$ the omitted mass.
For $0<\varepsilon_h<1$, let $o_h^{\mathrm O}$ be the normalized output of
the omitted keys. Then
\begin{equation}
\begin{aligned}
  o_h^{\mathrm D}-o_h^U
    &= \varepsilon_h(o_h^{\mathrm O}-o_h^U),\\
  \|o_h^{\mathrm D}-o_h^U\|
    &\le \varepsilon_h\operatorname{diam}\{v_j^{\mathrm e}:j<L\}.
\end{aligned}
\label{eq:jqs-selection-error}
\end{equation}
The diameter is the largest pairwise distance between effective values;
the error is zero at $\varepsilon_h=0$. With exact masses and arithmetic,
the tail policy ensures $\varepsilon_h\le\tau_h$ when its cap does not bind,
i.e., $\theta_h^\tau\ge\theta_h^c$. The group union can only reduce omitted
mass. A binding cap removes this guarantee, and fixed $N$ provides no such
bound. FP32 mass estimation perturbs the ideal guarantee; quantization error
relative to BF16 attention is a separate source of error.

\paragraph{Higher precision for the weighted sum.}
The gather kernel multiplies FP16 unnormalized softmax weights by FP16
reconstructed values, accumulating in FP32. This preserves higher precision
in the output sum while sparsity limits the number of values it processes.
Reconstruction retains NVFP4 storage error; it cannot restore BF16 values.

The precision choice also matches the cache layout: NVFP4 scales run along
the head dimension, the reduction dimension of QK, whereas PV reduces over
tokens. The gather kernel reconstructs selected values as it consumes them.
A separate kernel processes selected BF16 pages with FP16 operands.
The kernels emit partial outputs and softmax statistics for
a stable merge, followed by a BF16 output cast. The denominator uses FP32
weights; rounding the numerator weights to FP16 and the remaining kernel
arithmetic introduce error beyond Equation~\ref{eq:jqs-selected-output}.

\subsection{Execution and Prefill}
\label{sec:method-integration}

\paragraph{Realizing the decode savings.}
The implementation uses TensorRT-LLM's paged cache manager, CuTe~\citep{cecka2026cute} decode
kernels, and Triton~\citep{tillet2019triton} selection and BF16-page kernels. We store K scales
once in the layout consumed by the block-scaled MMA. Pure decode steps fuse
query packing with the new token's K/V store. Block lists remain on the
device, and both selection policies execute within CUDA graphs.

A sparse step reads every key in the scan, then rereads selected keys and
fetches their values. In a streaming model without cache reuse, the K/V-read
volume relative to dense NVFP4 attention approaches $\tfrac12+\kappa$, where
$\kappa$ is the selected block fraction. This approximation excludes
block-statistic traffic, fixed BF16-page overheads, and page padding.
It captures read savings when a small fraction is selected at long context
and extra reads when most blocks are selected. The dense NVFP4 control
directly attends all blocks with the same
cache and precision choices, omitting the scan and selector to isolate the
cost and benefit of selection.

\paragraph{Separating prefill from decode storage.}
Prefill uses FlashInfer's FP8 context attention with calibrated per-layer
power-of-two scales. Before attention, the store packs the persistent cache
directly from BF16 K/V, updates the two BF16 pages, and writes a temporary FP8
arena. Decode therefore incurs no FP8-to-NVFP4 conversion. We release a request's
arena pages when prefill finishes; decode reads only the NVFP4 pool
and BF16 side pages. Requests without arena space fall back to context
attention over expanded cache operands. Cache-capacity measurements must
account for the arena's device-memory reservation and the BF16 buffers
allocated across request slots.

\section{Evaluation}
\label{sec:experiments}

\subsection{Experimental Setup}
\label{sec:exp-setup}

\providecommand{\toprule}{\hline}
\providecommand{\midrule}{\hline}
\providecommand{\bottomrule}{\hline}
\providecommand{\expms}[2]{#1$\pm$#2}
\definecolor{exploss}{HTML}{9C3A2E}
\providecommand{\explow}[2]{\textcolor{exploss}{#1$\pm$#2\rlap{\smash{$^{\downarrow}$}}}}
\providecommand{\expup}[2]{#1$\pm$#2\rlap{\smash{$^{\uparrow}$}}}
\providecommand{\expmk}[1]{\rlap{\smash{$^{#1}$}}}

\begin{table*}[t]
\caption{Four-model comparison. Accuracy is mean $\pm$ SD over runs (\%).
BF16 has six runs on GPQA-D, AIME25, and LCB, and three on TB and SWE;
other rows have three runs. Red entries with $^{\downarrow}$ mark negative
paired differences from BF16 at $p<0.05$; $^{\uparrow}$ marks positive
differences at $p<0.05$,
not claimed as gains.
Decode is TPOT speedup at 128K context, batch 16, and 1,024 output tokens,
measured separately from accuracy; bold marks the fastest fitting method.
n/f: the batch exceeds the method's KV-pool capacity.
KV is packed bytes per token relative to BF16, including
format metadata but excluding serving reservations. $^{\S}$TriAttention
does not evict before 4,224 generated tokens; $^{\ddagger}$per retained token.}
\label{tab:exp-main}
\vskip 0.1in
\centering
\footnotesize
\setlength{\tabcolsep}{3pt}
\renewcommand{\arraystretch}{1.0}
\begin{tabular}{llrrrrrrr}
\toprule\noalign{\vskip 1pt}
Model & Method & GPQA-D & AIME25 & LCB v6 & TB 2.1 & SWE-V & Decode$\times$ & KV$\times$\\
\midrule\noalign{\vskip 1pt}
Qwen3.8 & BF16 & \expms{90.1}{0.9} & \expms{99.0}{0.4} & \expms{86.7}{0.8} & \expms{70.0}{2.6} & \expms{82.7}{3.7} & 1.00 & 1.000\\
27B & Jet-QSparse $N{=}64$ & \expms{90.9}{1.7} & \expms{98.3}{0.3} & \expup{88.5}{1.7} & \expms{71.9}{1.1} & \expms{83.5}{4.3} & \textbf{1.94} & 0.289\\
 & Quest & \expms{89.1}{1.1} & \expms{98.1}{0.4} & \explow{81.5}{0.4} & \explow{27.7}{3.6} & \explow{4.7}{0.2} & 1.68 & 1.055\\
 & SparQ & \expms{89.2}{1.1} & \expms{98.7}{0.4} & \expms{87.6}{0.6} & \expms{72.3}{1.7} & \expms{83.7}{2.9} & n/f & 1.501\\
 & TriAttention & \expms{88.4}{0.5} & \explow{79.0}{1.8} & \explow{66.4}{0.8} & \expms{73.0}{5.6} & \expms{82.7}{2.3} & 1.00\expmk{\S} & 1.000\expmk{\ddagger}\\
\midrule\noalign{\vskip 1pt}
Qwen3.6 & BF16 & \expms{84.8}{1.4} & \expms{91.8}{0.6} & \expms{79.5}{1.5} & \expms{41.6}{4.1} & \expms{68.5}{1.4} & 1.00 & 1.000\\
35B-A3B & Jet-QSparse $N{=}64$ & \expms{83.3}{2.2} & \expms{91.2}{0.6} & \expms{78.5}{0.3} & \expms{41.2}{1.7} & \expup{70.6}{1.6} & \textbf{1.57} & 0.289\\
 & Quest & \explow{82.0}{2.0} & \explow{76.1}{2.1} & \explow{59.6}{1.0} & \explow{22.1}{0.6} & \explow{19.4}{0.8} & 1.32 & 1.050\\
 & SparQ & \expms{83.0}{0.6} & \expms{91.1}{0.3} & \expms{77.7}{0.8} & \expms{40.1}{1.7} & \expms{69.3}{0.5} & 1.09 & 1.502\\
 & TriAttention & \explow{81.1}{1.1} & \explow{63.7}{1.3} & \explow{72.8}{0.6} & \expms{41.9}{2.3} & \expms{69.3}{1.4} & 1.00\expmk{\S} & 1.000\expmk{\ddagger}\\
\midrule\noalign{\vskip 1pt}
Nemotron 3 & BF16 & \expms{72.5}{2.2} & \expms{88.9}{1.0} & \expms{69.6}{0.6} & \expms{4.1}{1.7} & \expms{21.9}{1.8} & 1.00 & 1.000\\
Nano & Jet-QSparse $N{=}64$ & \expms{72.7}{1.8} & \explow{86.9}{0.5} & \expms{68.3}{1.4} & \expms{5.2}{1.7} & \expms{20.7}{1.7} & \textbf{1.11} & 0.297\\
 & Quest & \expms{72.4}{1.9} & \explow{85.5}{0.8} & \explow{67.4}{0.6} & \expms{4.9}{1.7} & \explow{19.3}{1.7} & 1.07 & 1.042\\
 & SparQ & \expms{71.2}{1.7} & \expms{87.2}{0.4} & \expms{68.6}{0.3} & \expms{2.6}{1.7} & \explow{17.7}{1.3} & 0.99 & 1.504\\
 & TriAttention & \expms{70.7}{0.9} & \explow{71.8}{1.0} & \explow{59.1}{1.3} & \expms{7.1}{1.3} & \expms{21.6}{2.1} & 1.00\expmk{\S} & 1.000\expmk{\ddagger}\\
\midrule\noalign{\vskip 1pt}
Nemotron 3.5 & BF16 & \expms{76.3}{2.0} & \expms{87.6}{0.6} & \expms{79.8}{0.9} & \expms{24.0}{0.6} & \expms{48.4}{0.2} & 1.00 & 1.000\\
Lightning & Jet-QSparse $N{=}64$ & \expms{77.6}{0.8} & \expms{86.6}{1.7} & \expms{79.1}{0.4} & \expms{24.3}{1.7} & \expms{48.8}{0.9} & \textbf{1.13} & 0.297\\
 & Quest & \expms{74.4}{0.8} & \expms{85.8}{1.9} & \explow{77.5}{0.1} & \expms{20.6}{1.7} & \explow{43.1}{1.0} & 1.05 & 1.042\\
 & SparQ & \expms{75.1}{1.1} & \expms{88.2}{0.3} & \expms{79.0}{0.3} & \expms{21.3}{4.1} & \explow{38.2}{1.0} & 0.97 & 1.504\\
 & TriAttention & \explow{72.9}{1.8} & \explow{72.7}{1.2} & \explow{69.4}{0.7} & \expms{23.2}{3.9} & \expms{47.8}{2.1} & 1.00\expmk{\S} & 1.000\expmk{\ddagger}\\
\bottomrule
\end{tabular}
\vskip -0.1in
\end{table*}

\textbf{Models and tasks.}
We evaluate Qwen3.8-27B, Qwen3.6-35B-A3B, Nemotron~3 Nano, and Nemotron~3.5
Lightning, changing only full-attention layers. We measure accuracy on GPQA-Diamond
\citep{rein2024gpqa}, AIME25 \citep{maa2025aime}, LiveCodeBench v6
\citep{jain2025livecodebench}, Terminal-Bench~2.1
\citep{tbench2025terminalbench}, and SWE-bench Verified
\citep{jimenez2024swebench,openai2024swebenchverified}, with three seeds per method.

\textbf{Implementation and baselines.}
The four-model evaluation uses frozen TensorRT-LLM snapshots on GB300,
with tensor parallelism one and a 256K context window. All methods share
the harness and seeds. We compare dense BF16 attention with Quest
\citep{tang2024quest}, SparQ \citep{ribar2024sparq}, and TriAttention
\citep{mao2026triattention}. Their aggregation budgets are 2,048 tokens,
the top 1/16 of positions, and the prompt plus 4,096 generated tokens,
respectively. On the Nemotrons, whose attention layers use no positional
encoding, our TriAttention integration uses identity phase factors, leaving
a calibrated mean-query dot product and a bandwise norm correction.
Each baseline uses one untuned operating point. For Jet-QSparse,
we evaluate dense NVFP4, two tail-mass settings, and $N=32,64,128$.
We selected $N=64$ for the main comparison after evaluating these settings.
These baseline re-implementations use Triton; Jet-QSparse uses CuTe decode
kernels alongside Triton selection and BF16-page kernels, so timings also
reflect implementation choices. The systems sweep also
evaluates TensorRT-LLM's FP8 KV cache for speed and capacity
\citep{trtllm_quantization}.

\textbf{Measurements.}
Accuracy and speed use separate workloads. Speedups are ratios of BF16 to
method decode time per output token (TPOT), measured on one GB300 per arm
with real text and CUDA graphs. The Qwen sweep uses two replicates,
32K--252K prompts, batches 1--64, and 4,096 generated tokens; we omit cells that
exceed an arm's measured pool. Output throughput includes prefill;
time to first token (TTFT) is separate. KV capacity counts allocated token
slots; bytes per token describe packed storage.

We compare accuracy using paired tests clustered by problem or task.
\emph{No detected loss} means that no task in the stated benchmark set has
a negative difference from BF16 at $p<0.05$; it does not establish equivalence.
We also report Benjamini--Hochberg (BH) adjusted $q$ values
\citep{benjamini1995fdr}, calling a negative result at $p<0.05$ but
$q\ge0.10$ a nominal loss. Approximate minimum detectable differences at
80\% power (MDD$_{80}$) describe resolution in percentage points.
Appendix~\ref{sec:exp-protocol} gives the tests, correction families,
run settings, and calibration checks. We exclude a separate SWE-bench
comparison after a shared writable environment exposed solution artifacts
across runs (Appendix~\ref{sec:exp-protocol}).

\begin{figure*}[t]
\centering
\setlength{\subfigtopskip}{0pt}
\setlength{\subfigcapskip}{1pt}
\setlength{\subfigbottomskip}{0pt}
\includegraphics{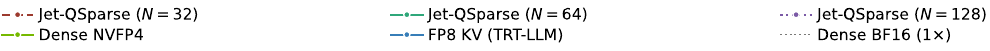}\par\vspace{3pt}
\subfigure[Qwen3.8, batch\label{fig:exp-bl-batch38}]{%
  \includegraphics{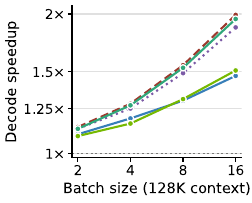}}\hfill
\subfigure[Qwen3.6, batch\label{fig:exp-bl-batch36}]{%
  \includegraphics{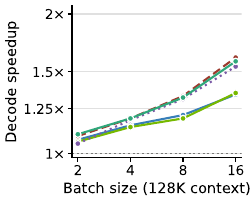}}\hfill
\subfigure[Qwen3.8, context\label{fig:exp-bl-context38}]{%
  \includegraphics{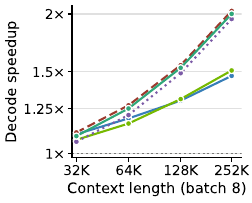}}\hfill
\subfigure[Qwen3.6, context\label{fig:exp-bl-context36}]{%
  \includegraphics{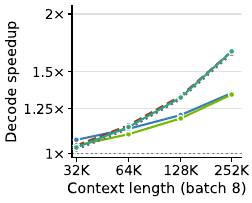}}
\caption{Jet-QSparse scaling with batch size and context length, compared
with dense NVFP4, FP8 KV, and dense BF16 on one GB300. Speedups use
4,096 output tokens and two replicates. (a,b) Batch sweeps from 2 to 16 at 128K context;
(c,d) context sweeps at batch eight. Jet-QSparse uses $N=32,64,128$.
Axes are logarithmic; we plot batches that fit BF16.
Each method uses its own BF16 timing reference; Appendix~\ref{sec:exp-system-details}
describes the comparison across measurements.}
\label{fig:exp-baselines}
\label{fig:exp-speed}
\end{figure*}

\subsection{Accuracy--Efficiency Trade-off}
\label{sec:exp-main}

At $N=64$, Jet-QSparse combines faster decode and compact storage with no
detected accuracy loss on the two Qwens and Lightning
(Table~\ref{tab:exp-main}). Across these models, MDD$_{80}$ ranges from
1.17--3.76 points on reasoning and coding, 2.44--2.80 on SWE-bench, and
6.24--8.58 on Terminal-Bench. Nano has one nominal loss: AIME25 decreases
by 1.98 points (95\% interval $[-3.77,-0.37]$, $p=0.0088$, $q=0.165$).
Increasing the budget to $N=128$ gives no detected loss on the five
application benchmarks across all four models. Section~\ref{sec:exp-ablation}
analyzes precision and budgets.

At 128K context and batch 16, generating 1,024 tokens, $N=64$ speeds up Qwen decode
by $1.94\times$ and $1.57\times$, and Nemotron decode by
$1.11\times$ and $1.13\times$. Its packed cache uses
28.9--29.7\% of BF16's bytes per token. Median contexts in the accuracy
runs are 6.9K--39.2K on reasoning tasks and 30K--65K per agentic call,
below 128K.

The baselines make different trade-offs (Table~\ref{tab:exp-main}).
SparQ has no detected loss on the Qwens, while its
layout uses $1.5\times$ the BF16 bytes. Quest incurs large agentic losses on the Qwens;
TriAttention loses long-reasoning accuracy. Jet-QSparse outpaces every
fitting baseline in Table~\ref{tab:exp-main} while using less KV storage;
SparQ's Qwen3.8 pool cannot admit this batch.

\subsection{Decode Performance and Serving Capacity}
\label{sec:exp-speed}

\textbf{Gains grow with batch and context.}
At $N=64$, decode speedup rises from 0.95--1.14$\times$ at batch one to
$2.00\times$ on Qwen3.8 and $2.01\times$ on Qwen3.6 at 252K, with batches
8 and 16, respectively (Appendix~\ref{sec:exp-system-details}). At BF16's
largest supported batches, $N=64$ outpaces every fitting baseline in
all eight tests (Table~\ref{tab:exp-bl}). Figure~\ref{fig:exp-baselines} compares scaling across block
budgets and dense precision baselines; at fixed batch eight, the 252K gains
are $2.00\times$/$1.66\times$.

\textbf{Native low-precision baselines.}
FP8 KV reaches $1.47\times$/$1.48\times$ decode speedup on the Qwens and
often matches or exceeds $N=64$ at batch one. At these two 252K workloads,
dense NVFP4 achieves $1.51\times$/$1.56\times$; selection adds
approximately $1.33\times$/$1.30\times$. The dense control captures the
cache format, arithmetic, and kernels together. The read model approaches
$\tfrac12+\kappa$ of dense NVFP4 traffic, so the scan limits gains from
smaller budgets.

\begin{figure}[!htbp]
\centering
\includegraphics[width=\columnwidth]{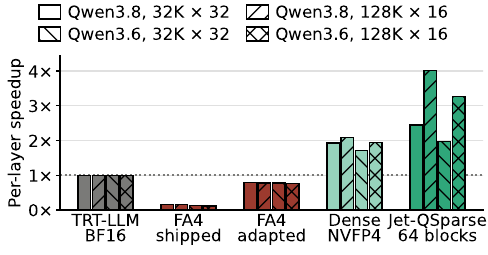}
\caption{GB300 single-layer decode speedup over TensorRT-LLM BF16 at head
dimension 256. Adapted FA4 uses GQA packing with or without split-KV,
whichever is faster. Jet-QSparse is timed separately, including BF16-page
work and selection for $N=64$.}
\label{fig:exp-kernel}
\label{fig:exp-kernel-cells}
\end{figure}

\textbf{Kernel-level speedup.}
Separate GB300 per-layer benchmarks at head dimension 256 give $N=64$
1.98--4.01$\times$ the served BF16 kernel's speed, including selection and
BF16-page work (Figure~\ref{fig:exp-kernel}). Appendix~\ref{sec:exp-kernel-details}
gives the timing protocol, detailed measurements, and FA4 baseline checks.

\textbf{Long generation.}
With 1K prompts, 16K outputs, and batch 64, $N=64$ runs at
0.97--1.08$\times$ BF16 speed across the four models: dense scanning and
selection offset gather savings at short contexts. TriAttention performs no
eviction in the 4K-output sweep; with 16K outputs it reaches
$1.16\times$/$1.06\times$ on the Qwens with eviction kernels compiled before
timing. Appendix~\ref{sec:exp-system-details} gives the full comparison.

\textbf{Capacity and serving throughput.}
Measured NVFP4 pools grow
2.34M$\to$6.05M tokens ($2.58\times$) on Qwen3.8 and
6.82M$\to$18.80M ($2.76\times$) on Qwen3.6; FP8 KV provides
$1.90\times$/$1.85\times$ BF16 capacity. Recurrent-state and FP8-arena
reservations keep NVFP4 below its $3.46\times$ packed-row ideal.
Qwen workloads exceeding BF16 capacity with 48 or 64
requests (128K prompts, 8K outputs) gain 1.99--2.16$\times$ BF16 throughput
from faster computation and larger admitted batches; both Qwen3.8 pools
overflow. The separate 4K-output sweep's two 252K workloads yield
$1.57\times$/$1.71\times$ throughput including prefill. Across all four models,
FP8 prefill improves single-request TTFT 1.17--1.32$\times$ at 128K and
1.25--1.46$\times$ at~250K.

\subsection{Design Ablations and Error Analysis}
\label{sec:exp-ablation}
\label{sec:exp-early}

\begin{figure*}[t]
\centering
\setlength{\subfigtopskip}{0pt}
\setlength{\subfigcapskip}{1pt}
\setlength{\subfigbottomskip}{0pt}
\includegraphics{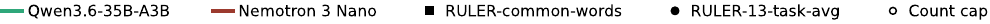}\par\vspace{3pt}
\subfigure[RULER accuracy\label{fig:pre-ladder-ruler}]{%
  \includegraphics{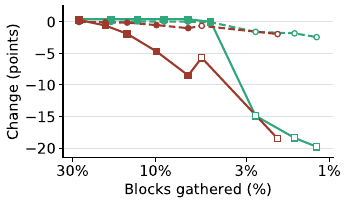}}\hfill
\subfigure[SWE-bench accuracy\label{fig:pre-ladder-swe}]{%
  \includegraphics{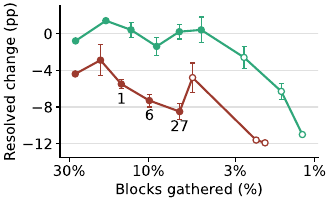}}\hfill
\subfigure[SWE-bench unfinished\label{fig:pre-ladder-unfinished}]{%
  \includegraphics{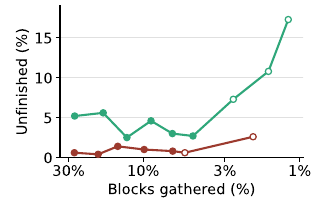}}
\caption{Precursor sparsity study. All block fractions were measured during
SWE-bench agentic runs. Hollow markers denote count caps: mass-threshold
selection with an additional per-head block limit.
(a) RULER common-word and 13-task score changes from each model's dense
reference. The x-axis identifies configurations
by their SWE-bench fractions, not their RULER traffic.
(b) SWE-bench changes from BF16 (Qwen3.6) or same-stack dense attention
(Nano); bars show half the two-seed spread. Numbers count head/layer pairs
retaining almost no history.
(c) Unfinished rates use available Qwen seeds and Nano seed one.}
\label{fig:pre-ladder}
\end{figure*}

\par\noindent\begin{minipage}{\columnwidth}
\captionof{table}{Paired AIME25 changes from BF16 (points). Red $^{\downarrow}$:
$p<0.05$. Final adds centering, two BF16 pages, FP8 prefill, and CuTe.
Lower rows use dense NVFP4 with the original prefill.
No listed gain survives BH correction.}
\label{tab:exp-precision}
\vskip 0.1in
\centering
\footnotesize
\setlength{\tabcolsep}{4pt}
\begin{tabular}{lrr}
\hline
Configuration & Nano & Lightning\\
\hline
$N=64$, unfixed & \textcolor[HTML]{9C3A2E}{$-4.13^{\downarrow}$} & \textcolor[HTML]{9C3A2E}{$-3.16^{\downarrow}$}\\
$N=64$, final & \textcolor[HTML]{9C3A2E}{$-1.98^{\downarrow}$} & $-1.01$\\
\hline
Dense, no BF16 pages & \textcolor[HTML]{9C3A2E}{$-2.74^{\downarrow}$} & \textcolor[HTML]{9C3A2E}{$-2.33^{\downarrow}$}\\
Dense, two BF16 pages & \textcolor[HTML]{9C3A2E}{$-2.12^{\downarrow}$} & $+0.66$\\
\quad + key centering & \textcolor[HTML]{9C3A2E}{$-1.63^{\downarrow}$} & $-0.38$\\
\hline
\end{tabular}
\end{minipage}\par

\textbf{Precision and calibration.}
We screen precision choices offline and measure task accuracy
(Table~\ref{tab:exp-precision}). Across all 12 Nemotron budget pairs,
the recipe raises AIME25 by 1.46--2.50 points over matched controls,
though no individual gain survives BH correction. These comparisons change
centering, BF16 retention, prefill, and the Nemotron kernel implementation together.
Dense controls separate BF16 retention from centering; centering adds no
detected reasoning or coding gain over two pages. At $N=64$, all 20 comparisons
with dense NVFP4 show no detected loss.

{\clubpenalty=10000
\textbf{Quantization changes trajectories.}
Diagnostic runs use simulated NVFP4 on H100 or packed NVFP4 on GB200,
without BF16 pages or FP8 prefill and with different budget and agentic
settings (Appendix~\ref{sec:exp-protocol}). On 16 Qwen3.6 SWE contexts,
NVFP4 changes 3.45\% of teacher-forced top-1 tokens; greedy continuations
diverge after a median 2.5 steps. This cohort is separate from the agentic
outcome comparisons in Appendix~\ref{sec:exp-precursor-quant}.
Table~\ref{tab:exp-main} provides the matched final-engine evaluation of
the complete Jet-QSparse precision recipe, rather than quantization alone.
\par}

\begin{figure}[!htbp]
\centering
\includegraphics{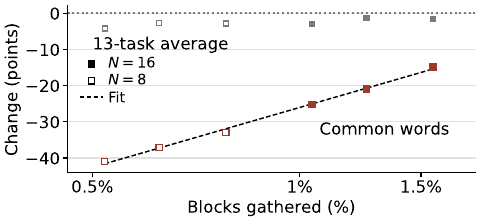}
\caption{Fixed-budget RULER study on Nemotron~3 Nano in the precursor
engine. Scores are changes from dense attention at the same context length
(128K, 160K, or 192K). Brick: common-word extraction; gray: 13-task average.
Filled markers use $N=16$ blocks per KV head; hollow markers use $N=8$.
The gathered fraction is $128N/L$. Dashed: fit to all six common-word points.}
\label{fig:pre-fixed}
\end{figure}

\textbf{Sparsity affects distributed evidence.}
Figure~\ref{fig:pre-ladder-ruler} compares the configurations on RULER,
indexed by their SWE-bench block fractions. Count caps balance attention
coverage against gather work: when a head needs more blocks to meet its
mass target, the cap takes precedence. The union across query heads can
still vary in size; fixed $N$ instead budgets the
shared KV-group list directly. Common-word losses reach
19.8 points for Qwen3.6 and 18.5 for Nano, while their 13-task averages
fall by 2.43 and 1.94 points. Needle retrieval remains largely preserved
across the tested budgets: all eight Qwen3.6 needle families score 100\%,
and Nano's needle scores stay within one point of dense.
Figure~\ref{fig:pre-fixed} shows aggregation losses with known fixed-budget
fractions. The selection-error relation
(Equation~\ref{eq:jqs-selection-error}) connects this sensitivity to omitted
mass and the gap between selected and omitted value averages.
Our final-engine RULER diagnostic \citep{hsieh2024ruler}
uses 13 tasks, three lengths (32K/128K/256K), and three 100-document datasets
per task and length. Table~\ref{tab:exp-ruler} reports its budget effects.
We report RULER as a separate diagnostic.

\textbf{Fixed budgets and context length.}
Figure~\ref{fig:pre-fixed} shows Nano's common-word loss increasing with
smaller gathered fractions, while the task average changes much less.
The six-cell fit gives 16.5 points per halving (SE 0.5), over 0.52--1.56\%.
For a binding budget, the per-step fraction is approximately
$\kappa=128N/L$. A practical starting rule is therefore to choose $N$ for
the configured maximum context and keep it fixed during decoding:
128 blocks at 256K and 512 at 1M both gather 6.25\% at maximum length.
The 1M setting remains untested; equal fractions need not retain equal
attention mass.

\textbf{Agentic failures differ by model.}
Figures~\ref{fig:pre-ladder-swe} and~\ref{fig:pre-ladder-unfinished} compare
accuracy and completion. Qwen3.6 stays near BF16 through $\kappa=0.048$;
tighter caps coincide with losses up to 11 points and 17.3\% unfinished
episodes, versus 3.6\% across preceding settings. Unfinished episodes yield
no gradable result after time/context limits or abnormal exits. Nano loses
5.6 points on average over comparable fractions; lower accuracy among
finished episodes accounts for at least 87\% of seed-one losses of
2.3 points or more. In the main evaluation, no Qwen3.6 SWE setting falls below
BF16 at $p<0.05$. Nano's two lowest-fraction losses remain nominal,
with no detected deficit against dense NVFP4.

{\clubpenalty=10000
\textbf{Budget choice and prefill precision.}
In the four-model study, $N=64$ retains at least 94\% of $N=32$ speed.
At $N=128$, all four models meet the five-benchmark accuracy criterion;
this budget covers 44--83\% of reference reasoning/coding contexts per step.
Small average RULER losses remain (Table~\ref{tab:exp-ruler}). A separate
FP8-prefill effect costs Qwen3.8 10.1--14.2 points on variable tracking
at 256K by echoing the answer prefix. Dense NVFP4 controls measure the
additional effect of sparsification; measuring task-level interaction with
quantization requires a matching BF16 sparse arm, which we did not evaluate.
\par}

\par\noindent\begin{minipage}{\columnwidth}
\captionof{table}{Final-engine RULER changes from BF16, averaged over 13 tasks and
three context lengths (points). Bold marks negative differences at
$p<0.05$ and $q<0.10$. Output caps affect Qwen3.8's positive aggregates,
concealing variable-tracking losses.}
\label{tab:exp-ruler}
\vskip 0.1in
\centering
\footnotesize
\setlength{\tabcolsep}{5pt}
\begin{tabular}{lrrrr}
\toprule
Budget & Qwen3.8 & Qwen3.6 & Nano & Lightning\\
\midrule
$N=32$ & $+0.12$ & $\mathbf{-0.85}$ & $\mathbf{-0.78}$ & $\mathbf{-1.55}$\\
$N=64$ & $+0.28$ & $\mathbf{-0.16}$ & $\mathbf{-0.45}$ & $\mathbf{-0.83}$\\
$N=128$ & $+0.28$ & $-0.02$ & $\mathbf{-0.29}$ & $\mathbf{-0.45}$\\
\bottomrule
\end{tabular}
\end{minipage}\par

\subsection{Scaling to a 2.4T-Parameter Model}
\label{sec:exp-scale}

We deploy Qwen3.8-2.4T-A95B \citep{qwen2026qwen38large} across eight
four-GPU GB300 nodes, with 32-way tensor and expert parallelism and
attention data parallelism.
All arms use the same FP8 weights; BF16 refers to the reference KV cache.
We evaluate GPQA-Diamond, AIME25, LiveCodeBench, and Terminal-Bench with
two seeds per Jet-QSparse arm. The reference has four runs on the first
three tasks and two on Terminal-Bench.

\par\noindent\begin{minipage}{\columnwidth}
\captionof{table}{2.4T accuracy (mean $\pm$ SD, \%). No arm falls below BF16
at $p<0.05$ on any of the four tasks.}
\label{tab:exp-scale}
\vskip 0.1in
\centering
\footnotesize
\setlength{\tabcolsep}{3pt}
\begin{tabular}{lrrrr}
\toprule
Method & GPQA-D & AIME25 & LCB & TB\\
\midrule
BF16 & \expms{92.2}{0.7} & \expms{99.3}{0.3} & \expms{89.2}{0.4} & \expms{82.6}{5.6}\\
Dense NVFP4 & \expms{91.7}{0.4} & \expms{99.6}{0.3} & \expms{89.2}{0.9} & \expms{81.5}{2.4}\\
$N=32$ & \expms{91.7}{0.4} & \expms{99.0}{0.3} & \expms{88.7}{0.5} & \expms{78.7}{1.6}\\
$N=64$ & \expms{92.4}{0.7} & \expms{99.0}{0.0} & \expms{89.6}{0.9} & \expms{79.8}{3.2}\\
$N=128$ & \expms{92.7}{1.1} & \expms{99.2}{0.0} & \expms{90.5}{0.0} & \expms{80.3}{0.8}\\
\bottomrule
\end{tabular}
\end{minipage}\par

All four configurations show no detected loss on the four evaluated tasks
(Table~\ref{tab:exp-scale}). At $N=64$, MDD$_{80}$ is 2.11 points on GPQA,
2.25 on LiveCodeBench, and 8.99 on Terminal-Bench. AIME25 is near its ceiling:
the paired change is $-0.36$ points with a 95\% interval of
$[-1.04,+0.29]$, and only five problems contribute nonzero paired differences.
We did not evaluate SWE-bench at this scale.

At $N=64$, decode speedups range from 1.07--1.22$\times$ across 32K--250K
workloads (Table~\ref{tab:exp-scale-speed}). The measured pool holds
2,360,832 tokens per rank versus 835,584
for BF16 ($2.825\times$). Twelve requests per rank, each with a 128K prompt
and 8K output, exceed BF16 capacity twofold but fit Jet-QSparse.
Whole-deployment throughput, including prefill, improves $1.487\times$
versus $1.423\times$ for dense NVFP4.

\par\noindent\begin{minipage}{\columnwidth}
\captionof{table}{2.4T speedups over dense BF16 on 32 GB300 GPUs. Decode columns
give prompt length $\times$ requests per rank. Output rate covers 12
requests per rank, each with a 128K prompt and 8K output, and includes
prefill. Ratios use two-run means, except $N=32,64$ use three-run medians;
their 128K replicate~ranges~overlap.}
\label{tab:exp-scale-speed}
\vskip 0.1in
\centering
\footnotesize
\setlength{\tabcolsep}{3pt}
\begin{tabular}{lrrrr}
\toprule
 & \multicolumn{3}{c}{Decode ($\times$)} & Output\\
Method & 32K$\times$23 & 128K$\times$6 & 250K$\times$3 & rate ($\times$)\\
\midrule
Dense BF16 & 1.000 & 1.000 & 1.000 & 1.000\\
Dense NVFP4 & 1.058 & 1.161 & 1.094 & 1.423\\
$N=32$ & 1.076 & 1.297 & 1.177 & 1.480\\
$N=64$ & 1.074 & 1.216 & 1.149 & 1.487\\
$N=128$ & 1.046 & 1.234 & 1.154 & 1.472\\
\bottomrule
\end{tabular}
\end{minipage}\par

\textbf{Attention share and memory.}
In the $N=32$ profiling study, we infer from traced kernel times that
attention accounts for 15--21\% of the BF16 step.
Reducing the NVFP4 pool to approximately BF16 capacity frees 41.5~GiB per
GPU. With large preallocated pools, the benefit instead appears as capacity;
in a replayed Terminal-Bench load, the reserved FP8 arena offsets live-KV
savings. Appendix~\ref{sec:exp-system-details} separates allocated memory
from live demand and describes these measurements.

\section{Conclusion}
\label{sec:conclusion}

Jet-QSparse combines compact NVFP4 storage and native QK computation with
sparse, higher-precision value aggregation in a simple, training-free design.
Its block layout aligns selection with contiguous memory access and
tensor-core execution. With 64 blocks per KV head, both Qwen models show
no detected accuracy loss on five application benchmarks and reach up to
$2.01\times$ end-to-end decode speedup over dense BF16 on separate 252K-context
workloads. Their measured KV pools hold $2.58$--$2.76\times$ as many tokens.
At 128 blocks, all four Qwen and Nemotron models show no detected accuracy
loss across mathematical reasoning, coding, and agentic benchmarks.
At 2.4T parameters, compact KV storage expands capacity while expert
computation and communication limit decode gains.

Quantization can change the attention distribution used for block selection.
The additional selection error depends on omitted mass and the
difference between selected and omitted value averages. Across all four
models, the 64-block configuration shows no detected additional loss
relative to the final dense NVFP4 control on the five application benchmarks.
Our diagnostics show larger losses on aggregation over distributed evidence
under aggressive sparsity, while needle retrieval remains largely preserved.
These findings guide joint
precision and budget calibration through attention-output probes and
end-to-end task validation.

\bibliography{references}
\bibliographystyle{mlsys2025}

\section*{AI Usage Disclosure}
Claude (under Claude Code) is used for kernel debugging (mainly tracing runtime errors and portability issues), experiment monitoring, and related work lookup.
ChatGPT (website) is used for wording polishing of the paper.

\clearpage
\appendix
\section{Evaluation Protocol and Statistical Resolution}
\label{sec:exp-protocol}

\paragraph{Runs and comparisons.}
The four-model accuracy study uses pinned checkpoints and frozen TensorRT-LLM
snapshots: \texttt{fsnap-10020414} for the final Jet-QSparse arms and
\texttt{final1001} for the BF16 reference and baseline reruns. Within each model
and benchmark, arms share the harness configuration, sampling settings,
262,144-token context limit, and strict tool-call parsers. Each arm runs seeds
42--44. On GPQA, AIME25, and LiveCodeBench, the reference averages two independent
BF16 runs per seed; Terminal-Bench and SWE-bench have one BF16 run per seed.
Tables report means and sample standard deviations across runs, whereas the
paired tests operate on item outcomes. We measure speed on separate workloads;
the timing cells do not measure latency within these accuracy runs.

Full-attention layers account for 16 of 64 layers on Qwen3.8-27B, 10 of 40
on Qwen3.6-35B-A3B, and six of 52 on each Nemotron. The Qwens use head
dimension 256 and the Nemotrons 128. The 252K sweep prompts use 258,048
of 262,144 tokens, leaving 4,096 for output.

Each seed evaluates 198 GPQA questions, 30 AIME25 problems with 16 samples
each, 454 LiveCodeBench v6 problems from the August 2024 to May 2025
evaluation window, 89 Terminal-Bench tasks, and 500 SWE-bench Verified instances.
Each agent call generates at most 16,384 tokens.

We chose the headline $N=64$ after evaluating all six configurations. We
selected the accuracy fixes on the same benchmark problems later used for
evaluation; repeat runs provide new samples, not new problems. Each sparse
baseline uses one untuned operating point, and the comparisons are not
sparsity-matched. These choices limit conclusions about an optimal budget or
the relative accuracy of independently tuned methods.

\paragraph{Timing and capacity eligibility.}
The common timing workload in Table~\ref{tab:exp-main} uses 131,072 prompt
tokens, batch size 16, and 1,024 output tokens on all four models.
Figure~\ref{fig:teaser} uses the same prompt length, batch size eight,
and 4,096 output tokens from the separate Qwen sweep. Speed runs use real
text and CUDA graphs on one GB300 per arm. TPOT is the per-token decode
time of the request prefilled last; speedup divides the same run's dense
BF16 TPOT by the arm's. SparQ's Qwen3.8 cell in Table~\ref{tab:exp-main}
is marked n/f: its 2,113,536-token demand exceeds the measured
1,576,928-token pool.

The batch/context sweeps use 4,096 output tokens and two replicates.
Baselines and Jet-QSparse run in separate campaigns, each normalized to
its own dense BF16 anchor. A same-node $N=32$ comparison reproduces the
Jet-QSparse grid with median absolute deviations of 0.08\% on Qwen3.8 and
0.85\% on Qwen3.6 (maxima 0.83\% and 2.73\%). This check supports the
comparison but does not eliminate cross-campaign uncertainty. Reported
fixed-batch cells require all requests to fit without suspension. We use
measured pool capacities where available and the validated cache-manager
allocation rule otherwise. We mark larger batches as not fitting and exclude
the Qwen3.8 FP8-KV 64K$\times$64 cell after a request suspension.

\paragraph{SWE evaluation integrity.}
We exclude a thirteen-arm Qwen3.8 SWE comparison over all 500 instances
and seeds 42--44 because shared writable runtime files compromised task
isolation. The sandboxes shared a Python installation containing a task
solution and notes.
Twenty retained episodes across ten arms and 19 tasks read these files;
one matching-task episode subsequently made the identical edit.
This establishes solution propagation, not its effect on the success rate.
Incomplete tool histories prevent bounding total exposure, so we exclude
the whole comparison rather than filter only known readers. The separate
FP8-KV/BLASST evaluation's BF16 control differs from
Table~\ref{tab:exp-main}'s reference by
$+4.83$ points (95\% CI $[+2.95,+6.71]$, $p=0.0001$).
We exclude the associated FP8-KV and BLASST SWE measurements from the
accuracy comparison. This audit cannot certify environments outside the
excluded comparison.

\paragraph{Paired, clustered inference.}
An item is a (seed, problem) pair on the reasoning benchmarks, with AIME25's
16 samples averaged within that item, or a (seed, task or instance) pair on the
agentic benchmarks. We exclude SWE-bench pairs with zero-token outcomes.
Let $d_i$ be the arm-minus-reference difference in percentage points. Items
cluster by problem, task, or instance across seeds. For cluster $c$, define
$D_c=\sum_{i\in c}d_i$, $n_c=|c|$, and $T=\sum_cD_c$. With $F=10{,}000$
seeded sign draws, independently uniform over $\{-1,+1\}$ for each cluster,
the estimate and two-sided Monte Carlo test are
\begin{equation}
\begin{aligned}
  \widehat\Delta &= \frac{\sum_c D_c}{\sum_c n_c},
  & T_f &= \sum_c s_{fc}D_c,\\
  p &= \frac{1+\#\{f:|T_f|\ge |T|\}}{F+1}.
\end{aligned}
\label{eq:exp-protocol-test}
\end{equation}
The 95\% interval inverts this test: it contains shifts $\delta_0$ for which
testing $D_c-\delta_0n_c$ gives $p\ge0.05$. This inversion assumes a constant
per-item shift. The sign-flip test is exact for independent cluster sums
symmetric about zero under the null, but approximate with a two-run reference.

\paragraph{Calibration and shared-reference uncertainty.}
Design-based grid simulations reject 3.3--4.7\% of nulls at nominal 5\%, with
94.2--96.2\% interval coverage in the simulated shifted cases. Real-data null
checks reject 0.0--3.3\% on three models but 4.4--13.3\% on Nano, indicating
that Nano's grid intervals are about 15\% too narrow. Agentic tests have
neither a null family nor a coverage simulation. All contrasts for a model
share its BF16 runs, so a high reference run can produce deficits across
several configurations. Clustering by task handles repeated tasks across
seeds but does not by itself remove whole-run shifts. Post hoc pooling and
sign analyses are descriptive; they do not determine configuration~verdicts.

\paragraph{Multiplicity and verdicts.}
For ordered $p$-values, Benjamini--Hochberg correction
\citep{benjamini1995fdr} gives
\begin{equation}
  q_{(i)}=\min_{j\ge i}\min\!\left(1,\frac{m p_{(j)}}{j}\right).
\end{equation}
The final-method family has $m=120$: four models, five benchmarks, and six
configurations comprising dense NVFP4, $N=32,64,128$, and $\rho=0.40,0.60$.
Baseline $q$-values use the full family of $m=80$ comparisons: the 60
displayed contrasts cover three baselines, four models, and five benchmarks;
20 additional comparisons remain in the correction family but are not displayed.
We apply multiplicity correction to RULER in separate diagnostic
families and exclude it from the five-benchmark criterion; we adopted this
separation after an initial combined analysis. A negative estimate with
$p<0.05$ is a detected loss if $q<0.10$, and nominal otherwise. A configuration
shows no detected loss only when none of its five cells is negative at
$p<0.05$, irrespective of the correction family. Positive differences are
not claimed as gains. The positive regression dependence condition for BH
\citep{benjamini2001control} is unproven for these two-sided tests; sharing a
reference does not establish it. We did not apply the Benjamini--Yekutieli
correction~\citep{benjamini2001control}.

\paragraph{Detection resolution.}
Using a normal approximation with interval half-width
$\eta\approx z_{0.975}\sigma$, we report
\begin{equation}
  \mathrm{MDD}_{80}=(z_{0.975}+z_{0.80})\sigma
  \approx1.429\eta.
\label{eq:exp-protocol-mdd}
\end{equation}
This approximates the loss detectable at $p<0.05$ with 80\% power, not an
equivalence margin or a confidence bound on loss. Multiplicity requires
stronger evidence: a lone rejection among 120 needs $p\le0.00083$ and has
only about 29\% power at one MDD$_{80}$. Discrete outcomes further limit
resolution. With $k$ nonzero cluster sums, the exact two-sided sign-flip
$p$ is at least $2/2^k$; fewer than six informative clusters cannot yield
$p<0.05$. Near-ceiling ties can also make the normal MDD approximation
optimistic. Smaller losses remain possible after a non-rejection.

\paragraph{The 2.4T extension.}
This campaign uses two seeds per Jet-QSparse arm, four BF16 grid runs, and two
BF16 Terminal-Bench runs, with one BH family of 16 contrasts across four
benchmarks. We deferred SWE-bench, so this study cannot meet the five-benchmark
criterion. Test and verdict rules preceded the runs. We chose two seeds
after reading two LiveCodeBench pilots, and chose the four-run reference
with BF16 grid scores known. We decided the final run set before Jet-QSparse
Terminal-Bench started, but registered it after reading the grid and one
BF16 Terminal-Bench run, 20 hours after the first Jet-QSparse Terminal-Bench
launch and before reading any of its rows. Two runs had started and one had
finished by registration. Under equal independent sampling variance, two seeds
increase MDD$_{80}$ by approximately $\sqrt{3/2}=1.22$ relative to three; each
cell nevertheless uses its estimated resolution.

\paragraph{Exploratory error studies.}
The precursor studies in Section~\ref{sec:exp-ablation} use a separate engine,
precision recipe, and agentic protocol. Appendix~\ref{sec:exp-precursor-quant}
details the heterogeneous quantization comparisons and their descriptive
pool. The token disagreement study uses 16 contexts from one model and is
not paired with those task-outcome comparisons. The context-scaling fit uses six single-run
cells with shared controls within each length and different documents across
lengths; its standard error does not represent fresh-document uncertainty.
We selected the agentic budget range from Qwen's outcomes; Nano's
finished-episode decomposition uses seed one. These observations suggest
patterns to investigate; the final-engine controls provide separate checks
in the main text.

\section{Kernel Measurements}
\label{sec:exp-kernel-details}

\begin{table}[ht]
\caption{Single-layer decode times in $\mu$s on GB300, head dimension
256. Qwen3.8 has 24 query/four KV heads; Qwen3.6 has 16/two.
JQS denotes Jet-QSparse, timed separately with BF16-page work and selection
for $N=64$.}
\label{tab:exp-kernel}
\vskip 0.1in
\centering
\footnotesize
\setlength{\tabcolsep}{2.5pt}
\begin{tabular}{lrrrr}
\toprule
 & \multicolumn{2}{c}{Qwen3.8} & \multicolumn{2}{c}{Qwen3.6}\\
Method & 32K$\times$32 & 128K$\times$16 & 32K$\times$32 & 128K$\times$16\\
\midrule
TRT-LLM BF16 & 580.6 & 1,135 & 291.5 & 571.7\\
FA4, as shipped & 3,515 & 7,499 & 2,195 & 4,934\\
FA4, adapted & 741.9 & 1,465 & 374.1 & 747.0\\
JQS, dense & 301.0 & 545.8 & 170.2 & 293.4\\
JQS, $N=64$ & 237.2 & 283.2 & 147.5 & 175.4\\
\bottomrule
\end{tabular}
\end{table}

\begin{figure}[t]
\centering
\setlength{\subfigtopskip}{0pt}
\setlength{\subfigcapskip}{1pt}
\setlength{\subfigbottomskip}{0pt}
\includegraphics{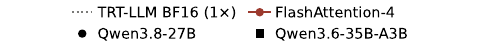}\par\vspace{3pt}
\subfigure[FA4 decode by batch\label{fig:exp-kernel-batch}]{%
  \includegraphics{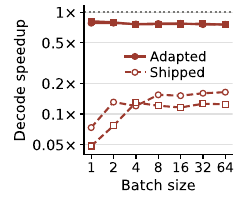}}\hfill
\subfigure[FA4 prefill\label{fig:exp-kernel-prefill}]{%
  \includegraphics{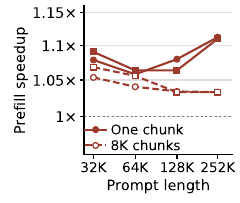}}
\caption{FA4 speedup over TRT-LLM BF16 (log scale). (a) Decode at 128K:
as shipped (hollow, dashed) or adapted (filled, solid).
(b) Prefill at batch one: one chunk
(filled, solid) or 8K chunks (hollow, dashed). Circles and squares denote
the Qwen3.8 and Qwen3.6 head geometries.}
\label{fig:exp-fa4}
\end{figure}

The TensorRT-LLM/FlashAttention-4 study uses FA4 4.0.0b19, CUDA-event
timings with L2 flushes, and two replicate medians with at least 20 timing
repetitions. Decode runs under CUDA graphs. The tested FA4 path at head
dimension 256 lacks native GQA packing and split-KV. The adapted result is the
faster of GQA packing alone and packing with an external split-KV/combine
step. We time Jet-QSparse's final decode kernels in a separate benchmark with
seven shuffled rounds and include scan, selection, sparse attention,
BF16-page operations, and merging where applicable. The descriptive ratios
in Figure~\ref{fig:exp-kernel} compare matching head geometries, batches, and
context lengths. Differences in harnesses, clocks, and timing aggregation
prevent an exact decomposition of end-to-end speed.

Figures~\ref{fig:exp-kernel} and~\ref{fig:exp-fa4} use TRT-LLM time divided by method time.
The decode reference is the BF16 attention kernel itself. The prefill
reference is the corresponding served context operator, using either one
chunk or 8K chunks on both sides. FA4 prefill takes 0.899--0.945 of the
one-chunk baseline time and 0.936--0.968 with 8K chunks. These are
attention measurements, not a separate FA4 end-to-end serving arm.

\section{Precursor Quantization Diagnostics}
\label{sec:exp-precursor-quant}

\begin{figure}[t]
\centering
\includegraphics{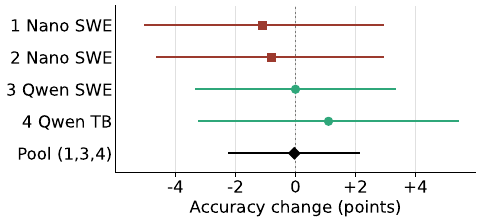}
\caption{Exploratory agentic accuracy changes from BF16 to NVFP4, with
95\% intervals. Study protocols differ; rows are not matched repetitions
across models. SWE denotes SWE-bench Verified; TB denotes
Terminal-Bench~2.1. The descriptive pool uses studies 1, 3, and 4;
study 2 shares study 1's BF16 reference.}
\label{fig:pre-quant}
\end{figure}

Figure~\ref{fig:pre-quant} summarizes four earlier agentic comparisons of
NVFP4 and BF16. These studies differ in hardware, arithmetic, benchmark,
and inclusion of incomplete episodes. They provide exploratory evidence
separate from the matched final-engine comparisons in
Table~\ref{tab:exp-main}.

\paragraph{Study protocols.}
Studies 1 and 2 use simulated NVFP4 with BF16 arithmetic on H100 for
Nemotron~3 Nano SWE-bench Verified. They share one BF16 reference and include only
episodes completed by both arms: 468 and 477 of 500 problems, respectively.
Study 3 evaluates Qwen3.6 on all 500 SWE-bench Verified problems using packed NVFP4
on GB200. Its records conflict on sampling settings and whether decode
uses native FP4 arithmetic or BF16 over a dequantized cache, limiting
attribution to quantization alone. Study 4 evaluates Qwen3.6 on 89
Terminal-Bench~2.1 tasks with four seeds per arm on GB200 and native FP4
scoring at prefill and decode.

\paragraph{Descriptive pooling.}
The inverse-variance pool of studies 1, 3, and 4 gives $-0.04$ points
[95\% CI $-2.2$, $+2.1$]. We exclude study 2 because it shares study 1's
BF16 reference. The pool assumes a common percentage-point effect across
these different models and benchmarks; it does not reconcile their
protocols or completion criteria. It establishes neither equivalence nor
the accuracy of the final Jet-QSparse recipe. The 16-context
probe in Section~\ref{sec:exp-ablation} uses a separate cohort and cannot
link token disagreement to task outcomes.

\section{Additional System Measurements}
\label{sec:exp-system-details}

\subsection{Batch, Context, and Block Budget}

Table~\ref{tab:exp-bl} compares baselines at the largest
batch dense BF16 accommodates at each context. It also reports each
method's best output rate, including prefill, over its tested batches.
This output rate reflects both faster execution and additional
capacity. Ratios use each campaign's own BF16 reference;
Appendix~\ref{sec:exp-protocol} describes the timing comparison and its
$N=32$ consistency check.

\begin{table*}[t]
\caption{Measured KV-pool capacity and performance on the Qwens. Pool is the
token capacity relative to dense BF16. TPOT columns use BF16's largest
batch at each context: 32/32/16/8 on Qwen3.8 and 64/64/32/16 on Qwen3.6.
Output-rate columns divide each method's best throughput, including prefill,
by BF16's best over the tested batches. All cells generate 4,096 tokens;
n/f means the batch does not fit the measured pool. The Qwen3.8 FP8-KV
64K$\times$64 cell is excluded after a request suspension; its cells beyond
BF16 capacity run with idle neighboring GPUs. Jet-QSparse and baseline
ratios come from separate campaigns.}
\label{tab:exp-bl}
\vskip 0.1in
\centering
\footnotesize
\setlength{\tabcolsep}{4pt}
\begin{tabular}{lrrrrrrrrr}
\toprule
 & Pool & \multicolumn{4}{c}{TPOT speedup} & \multicolumn{4}{c}{Best output-rate ratio}\\
Method & ratio & 32K & 64K & 128K & 252K & 32K & 64K & 128K & 252K\\
\midrule
\multicolumn{10}{l}{\emph{Qwen3.8-27B}}\\
Quest & 0.95 & 1.10 & 1.48 & 1.70 & 1.83 & 1.07 & 1.24 & 1.29 & 1.27\\
SparQ & 0.67 & 1.15 & n/f & n/f & n/f & 1.10 & 0.91 & 0.91 & 0.92\\
TriAttention & 1.00 & 1.00 & 1.00 & 1.00 & 1.00 & 1.00 & 1.00 & 1.01 & 1.01\\
FP8 KV & 1.90 & 1.30 & 1.46 & 1.47 & 1.47 & 1.50 & 1.28 & 1.50 & 1.55\\
Dense NVFP4 & 2.58 & 1.30 & 1.49 & 1.51 & 1.51 & 1.52 & 1.49 & 1.54 & 1.60\\
Jet-QSparse $N=64$ & 2.58 & 1.41 & 1.84 & 1.95 & 2.00 & 1.63 & 1.69 & 1.76 & 1.81\\
\midrule
\multicolumn{10}{l}{\emph{Qwen3.6-35B-A3B}}\\
Quest & 0.96 & 1.04 & 1.32 & 1.53 & 1.74 & 1.03 & 1.19 & 1.26 & 1.25\\
SparQ & 0.69 & 1.09 & 1.17 & 1.18 & 1.20 & 1.07 & 1.11 & 1.09 & 1.08\\
TriAttention & 1.00 & 1.01 & 1.00 & 1.01 & 1.00 & 1.01 & 1.00 & 1.01 & 1.01\\
FP8 KV & 1.85 & 1.25 & 1.38 & 1.43 & 1.48 & 1.21 & 1.31 & 1.57 & 1.58\\
Dense NVFP4 & 2.76 & 1.25 & 1.43 & 1.48 & 1.56 & 1.21 & 1.34 & 1.63 & 1.81\\
Jet-QSparse $N=64$ & 2.76 & 1.31 & 1.68 & 1.87 & 2.01 & 1.26 & 1.48 & 1.88 & 2.08\\
\bottomrule
\end{tabular}
\end{table*}

We compute capacity ratios from measured attention-pool sizes. The cache manager
divides its quota between attention pages and recurrent states using an
average request length of 16,384 tokens. Reducing bytes per attention token
therefore increases the recurrent-state allocation as well as token
capacity. This explains why FP8 KV provides 1.90/1.85$\times$ capacity and
NVFP4 2.58/2.76$\times$, below their storage-only ratios. SparQ's additional
key copy reduces its pool to 0.67/0.69 of BF16. The allocation rule matches
measured pool counts within 0.1\% and predicts approximately
$1.88\times$ NVFP4 capacity on the Nemotrons. We exclude fixed-batch
timings that require multiple admission waves.

\begin{figure}[t]
\centering
\includegraphics[width=\columnwidth]{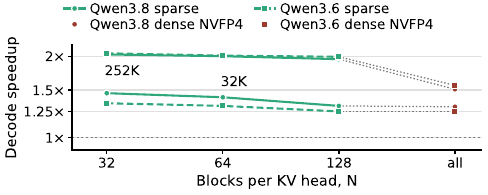}
\caption{Block-budget sweep on one GB300, with 4,096 output tokens and two
replicates. TPOT speedup is relative to dense BF16 at its largest fitting
batch: Qwen3.8 uses 32K$\times$32 and 252K$\times$8; Qwen3.6 uses
32K$\times$64 and 252K$\times$16. Circles and solid lines denote Qwen3.8;
squares and dashed lines denote Qwen3.6. Jade marks sparse budgets; brick
marks dense NVFP4 at ``all.''}
\label{fig:exp-speed-n}
\end{figure}

The sparse curves in Figure~\ref{fig:exp-speed-n} vary the
gather budget while retaining the dense scan. At BF16-fitting batches, $N=128$ is at most 12\% slower than
$N=32$ on the Qwens; the difference reaches 19\% at Qwen3.8's
32K$\times$64 cell beyond BF16 capacity. The scan and non-attention layers
limit the benefit of reducing the gather alone.

\subsection{Long Generation and Kernel Scope}

\begin{table}[t]
\caption{Long-generation speedups: 1,024-token prompts, 16,384 outputs,
and batch 64. Top: TPOT relative to dense BF16; two replicates, with
baselines and Jet-QSparse in separate campaigns.
$^{*}$Eviction kernels compiled before timing; two same-node GPU pairs
with iteration statistics enabled. $^{\dagger}$Rebased on other runs'
BF16 reference, with approximately 5\% uncertainty.
Bottom: one attention layer at 8.5K$\times$64 on random data, relative to
FlashInfer BF16; Jet-QSparse kernels omit the accuracy fixes.}
\label{tab:exp-longgen}
\vskip 0.1in
\centering
\footnotesize
\setlength{\tabcolsep}{3pt}
\begin{tabular}{lrrrr}
\toprule
 & Q3.8 & Q3.6 & Nano & Lightning\\
\midrule
\multicolumn{5}{l}{\emph{End to end}}\\
Quest & 0.69 & 0.81 & 0.86 & 0.91\\
SparQ & 1.02 & 0.97 & 0.97 & 0.93\\
TriAttention & $1.16^{*}$ & $1.06^{*}$ & 1.01 & 0.96\\
Dense NVFP4 & 1.14 & 1.07 & $1.00^{\dagger}$ & 0.99\\
Jet-QSparse $N=64$ & 1.08 & 1.01 & $0.98^{\dagger}$ & 0.97\\
Conditional limit & 1.43 & 1.19 & 1.08 & 1.06\\
\midrule
\multicolumn{5}{l}{\emph{One attention layer}}\\
Quest & 2.04 & 1.89 & \multicolumn{2}{c}{1.31}\\
SparQ & 1.16 & 0.88 & \multicolumn{2}{c}{0.74}\\
TriAttention & 1.54 & 1.48 & \multicolumn{2}{c}{1.40}\\
Dense NVFP4 & 1.66 & 1.49 & \multicolumn{2}{c}{0.94}\\
Jet-QSparse $N=64$ & 1.18 & 1.00 & \multicolumn{2}{c}{0.55}\\
\bottomrule
\end{tabular}
\end{table}

TriAttention first evicts after 4,224 generated tokens. With a 1K prompt,
its cache cycles between 5,120 and 5,248 tokens while dense attention
reaches 17,408. All 64 requests fit the BF16 pool in
Table~\ref{tab:exp-longgen}, so this experiment measures shorter reads,
not additional admitted requests. At 4,096 output tokens in the batch
sweep, no eviction occurs. Compiling the eviction kernels before timing
avoids charging a one-time compilation to decode. The Qwen measurements
use this warm-up; the Nemotron cells retain the standard timing protocol.
Same-method cross-campaign ratios differ by up to $0.05\times$ in the
Qwen3.8 long-generation cell, so small gaps across campaigns are descriptive.

\paragraph{Short-prompt KV memory.}
Figure~\ref{fig:teaser} also measures memory with eight requests, 4K prompts,
and 16K outputs on one GB300 per method. All requests finish without pauses.
This single-run check sets maximum batch size to 8, sequence length to
20,480, and prefill token budget to 8,192. We convert measured peak
used pages and page sizes to occupied bytes, including layout metadata,
and add the two BF16 pages for each active Jet-QSparse request. Ratios use
the matching engine's BF16 control; both dense controls have identical
attention-memory values. Occupied bytes exclude recurrent states, unused
reservations, and the FP8 prefill arena. The 128K/4K comparison instead uses
the two-run sweep with configured maxima of 64 requests and 262,144 tokens
for sequence length and the prefill token budget.

The conditional limits in Table~\ref{tab:exp-longgen} use $1/(1-a)$,
assuming non-attention time is unchanged, with estimated attention shares
of 30\%, 16\%, 7\%, and 6\%. These estimates scale the 8.5K single-layer
times by the number of attention layers; they approximate a generation
whose context changes throughout decoding. At these short contexts,
$N=64$ gathers a traffic-weighted block fraction of 0.72, so the read model
in Section~\ref{sec:method-integration} predicts
$\tfrac12+0.72=1.22$ times the dense NVFP4 K/V traffic.

Single-layer speed does not fully predict serving speed. Quest's layer
times, with its first two attention layers kept dense, imply
1.01--1.15$\times$ end-to-end speedup under an unchanged
non-attention remainder, but measured speed is 0.69--0.91$\times$. We did not
decompose this gap, which limits algorithm-level conclusions
from its serving~ratios.

\subsection{The 2.4T Decode Step and Memory Use}

\begin{table}[t]
\caption{2.4T systems study on 32 GPUs, using $N=32$. Step times and
speedups use TPOT estimates with traced windows replaced by untraced
means (two allocations). Attention times are rank-0 kernel attributions;
the BF16 attention share $a$ is inferred because profiling overhead
exceeds its tolerance. The limit $1/(1-a)$ assumes unchanged
non-attention time. Times are milliseconds.}
\label{tab:exp-r19}
\vskip 0.1in
\centering
\footnotesize
\setlength{\tabcolsep}{2pt}
\begin{tabular}{lrrrrrr}
\toprule
 & BF16 & \multicolumn{2}{c}{Attention} & & & $N=32$\\
Cell & step & BF16 & $N=32$ & $a$ & Limit & speedup\\
\midrule
32K$\times$23 & 69.6 & 10.2 & 4.0 & 14.7\% & 1.17 & 1.10\\
128K$\times$6 & 58.3 & 10.5 & 2.8 & 18.0\% & 1.22 & 1.20\\
250K$\times$3 & 47.0 & 9.8 & 2.7 & 20.9\% & 1.26 & 1.13\\
\bottomrule
\end{tabular}
\end{table}

Table~\ref{tab:exp-r19} attributes 15--21\% of the BF16 step to attention
and shows how $N=32$ reduces that component. This profiling study uses
separate runs from the deployment measurements in
Table~\ref{tab:exp-scale-speed}.
At 32K$\times$23, sampled traced kernels additionally attribute 20.0\,ms
to routed experts, 18.7\,ms to MoE communication including waits, and
14.9\,ms to BF16 GEMMs. These sampled-rank timings do not give an
untraced whole-system decomposition. The non-attention remainder also
varies between arms, so $1/(1-a)$ is a conditional limit rather than an
unconditional bound on speedup.

\paragraph{Allocated memory and live demand.}
With the pool using 85\% of free memory, measured NVML plateaus are
265.03\,GiB per GPU for dense BF16 and 262.74\,GiB for $N=32$; reservation
keeps these footprints nearly constant across loads. Reducing the NVFP4
pool fraction to 0.45 to approximate BF16's token capacity lowers its
plateau to 225.40\,GiB. Against the corresponding BF16 reference mean of
266.85\,GiB, this frees 41.45\,GiB per GPU, or approximately 1,327\,GiB
per 32-GPU copy. Propagating the bimodal BF16 reference's spread gives
an SD of 3.38\,GiB per GPU, despite agreement between the two NVFP4
replicates. Normalizing to exactly 835,584 tokens gives a calculated
42.76\,GiB saving. Associated TPOT changes are unresolved at
MDD$_{80}$ values of 1.29--5.37\,ms.

A replay of 1,113 Terminal-Bench calls measures demand under an agentic
arrival pattern. It reproduces token counts and timing, not task content,
and uses two paired measurements. Subtracting unused reserved KV space
from measured allocation gives peak memory needs of 200.97\,GiB for BF16
and 205.31\,GiB for $N=32$, a derived difference of
$+4.34\pm0.18$\,GiB. These are accounting estimates, not NVML footprints.
Peak live KV averages about 105K/94K tokens across the two pairs, well
below either pool. Hypothetically shrinking the FP8 arena to its observed peak
would change the difference to $-4.75$/$-4.49$\,GiB in the two pairs;
we did not measure that allocation policy. We amended the accounting filters
after a diagnostic measurement. In these replayed calls, decode
accounts for 97.8\% of client model time, with 68.7--69.3\% of decode
time in steps that also prefill another request.

\paragraph{Other execution choices.}
FP8 non-expert weights reduce the 32K$\times$23 step by 3.3--3.5\,ms.
The reduction is resolved for $N=32$ (MDD$_{80}$ 1.3\,ms), but not for
dense BF16 (4.2\,ms); the sparse-to-dense speedup change remains within
MDD$_{80}$ 0.07--0.14. Attention TP16 on four nodes achieves
8.9--9.7\,ms for one request and 250/319 output tokens per second per
node for dense BF16/NVFP4 at 128K$\times$30. We did not evaluate
fixed-budget selection on this layout because each KV group has four query heads
instead of sixteen. A low-$M$ GEMM dispatcher saves 1.5--2.4\,ms in
isolated step measurements. We evaluated these weight and execution variants
for performance only; they carry no accuracy claim.

\end{document}